\documentclass[10pt,twocolumn,letterpaper]{article}

\usepackage{cvpr}              

\definecolor{cvprblue}{rgb}{0.21,0.49,0.74}
\usepackage[pagebackref,breaklinks,colorlinks,allcolors=cvprblue]{hyperref}

\def\paperID{15669} 
\def\confName{CVPR}
\def\confYear{2025}

\title{Mesh Mamba: A Unified State Space Model for Saliency Prediction in Non-Textured and Textured Meshes}

\author{Kaiwei~Zhang\textsuperscript{\rm 1}, Dandan~Zhu\textsuperscript{\rm 2}$^*$, Xiongkuo~Min\textsuperscript{\rm 1}$^*$, Guangtao~Zhai\textsuperscript{\rm 1}\thanks{Corresponding author.}\\ 
\textsuperscript{\rm 1}Shanghai Jiao Tong University, \textsuperscript{\rm 2}East China Normal University\\
{\tt\small https://github.com/kaviezhang/MeshMamba}
}

\begin{document}
\maketitle
\begin{abstract}
Mesh saliency enhances the adaptability of 3D vision by identifying and emphasizing regions that naturally attract visual attention. To investigate the interaction between geometric structure and texture in shaping visual attention, we establish a comprehensive mesh saliency dataset, which is the first to systematically capture the differences in saliency distribution under both textured and non-textured visual conditions. Furthermore, we introduce mesh Mamba, a unified saliency prediction model based on a state space model (SSM), designed to adapt across various mesh types. Mesh Mamba effectively analyzes the geometric structure of the mesh while seamlessly incorporating texture features into the topological framework, ensuring coherence throughout appearance-enhanced modeling. More importantly, by subgraph embedding and a bidirectional SSM, the model enables global context modeling for both local geometry and texture, preserving the topological structure and improving the understanding of visual details and structural complexity. Through extensive theoretical and empirical validation, our model not only improves performance across various mesh types but also demonstrates high scalability and versatility, particularly through cross validations of various visual features.
\end{abstract}
    
\section{Introduction}
\label{sec:intro}

In 3D visual perception, geometric structure and texture information are critical factors in determining how the human eyes capture and interpret object features. Geometry conveys essential visual information through an object's contours, volume, and surface characteristics, facilitating rapid recognition of its overall structure and spatial arrangement. In contrast, texture enriches visual detail by simulating real-world appearance through surface patterns, color variations, and material qualities, thus elevating the realism of the visual experience.
Together, geometry and texture create a comprehensive visual representation that shapes visual attention distribution and efficiency in object recognition. A deeper understanding of the interplay between these factors sheds light on understanding how the human eye processes complex 3D environments, offering insights that can guide model design to improve the overall visual experience.

Advancements in saliency analysis rely on the integrated study of texture and geometric features, particularly within dynamic and realistic virtual environments. Virtual reality (VR) technology has revolutionized the field of 3D saliency collection by enabling precise tracking of visual attention within immersive virtual spaces, as evidenced by studies like \cite{ding2023towards,martin2024sal3d}. However, current research mainly focuses on non-textured meshes and simple vertex colors, with limitations in experimental design and dataset scale, which restrict the effectiveness and generalizability of the data collected.

To address these limitations, we design an innovative VR eye-tracking experiment and construct the first dataset to systematically capture saliency differences between textured and non-textured conditions, providing a comprehensive record of saliency distributions for the same 3D mesh model under varying visual contexts. Through an immersive six degrees of freedom (6-DOF) space, our experimental setup allows for high-precision eye-tracking data collection, enabling us to explore in depth how texture influences visual attention and the interaction between texture and geometry in shaping saliency.

Even more significantly, we propose mesh Mamba, a state space model \cite{gu2023mamba,dao2024transformers} for mesh learning that is capable of performing saliency prediction for both non-textured and textured meshes.
The graph convolution encoder begins with mapping the triangular mesh surfaces to UV pixel positions in the corresponding texture images. This mapping is transformed into continuous implicit representations \cite{sitzmann2020implicit,mildenhall2021nerf,zhang2022implicit} within a high-dimensional latent code domain of the texture, aligning 2D texture details precisely with 3D geometry.
Local geometric features, such as triangular face shapes, convex curvature, and spatial distribution, are integrated alongside texture information. The graph convolution then leverages mesh connectivity of adjacent faces to generate a topologically structured feature representation.

The Mamba block employs a specialized approach featuring subgraph embedding and a bidirectional SSM for global feature context modeling. Subgraphs are sampled around farthest point sampling (FPS) centers using random walk sampling (RWS), effectively segmenting the mesh into topology-preserving local patches. Through feature diffusion and integration, the receptive field of local features is expanded, capturing a richer context.
The bidirectional SSM further processes contextual information across the sequence, allowing a more comprehensive consideration of patch token positions and interrelationships within the global structure, thereby enhancing the model’s understanding of overall geometry and content.
In the final step, feature propagation employs voting interpolation to upscale the output sequence for dense predictions. This model effectively adapts to structural complexity and visual detail by leveraging both local and global mesh features, delivering reliable predictions across diverse 3D models.

Validation experiments are conducted on both non-textured and textured meshes, showing accurate predictions aligned with ground truth. Cross validations using geometry, color, and texture reveal how each visual feature influences saliency, confirming the model's generalizability across mesh types and its effectiveness in capturing visual attention patterns.
We summarize the key contributions as follows:

\begin{itemize}
    \item The mesh saliency dataset to systematically capture saliency differences across contrasting visual contexts, specifically under textured and non-textured conditions.
    \item Mesh Mamba, a state space model for saliency prediction, engineered to accommodate complex geometries and textural details, adaptable to diverse visual stimuli and perceptual cues.
    \item A novel subgraph embedding approach that mitigates topological disruption from spatial clustering, preserving inherent positional relationships and geometric integrity.
    \item A bidirectional SSM for global context modeling that enables cohesive interaction across token sequences, bridging local geometry with texture channels.
\end{itemize}

\section{Related Work}
\noindent
\textbf{Mesh Saliency Datasets and Methods.}
The concept of mesh saliency was first introduced by Lee et al. \cite{lee2005mesh}, who identified geometric shape as the primary determinant of saliency.
Research on mesh saliency informed resource allocation by focusing computational efforts on areas of interest, which optimized rendering performance in fields like graphics and game development. For instance, in game engines, Level of Detail (LOD) techniques \cite{heok2004review} adjusted mesh complexity based on viewing distance. Integrating LOD with mesh saliency allowed for targeted reductions in polygon density within non-salient regions, thus preserving essential details and visual experience \cite{nehme2023textured,zhang2023synergetic} in more critical regions.
Recently, a growing number of saliency datasets for 3D meshes have been introduced. In earlier studies \cite{chen2012schelling,giorgi2007shape,dutagaci2012evaluation}, mesh saliency was typically annotated using mouse tracking.
Studies such as \cite{kim2010mesh,lavoue2018visual} mapped the recorded 2D coordinates onto the 3D models using on-screen eye trackers to capture gaze points. In subsequent studies \cite{ding2023towards,martin2024sal3d,zhang2024textured}, eye-tracking data for 3D meshes were gathered within a VR environment.
Despite significant progress in developing 3D mesh saliency datasets, many limitations remain, particularly concerning dataset size and the lack of consideration for the impact of textured meshes.
As for mesh saliency prediction methods, 2D-to-3D mapping techniques \cite{lee2005mesh,abid2020towards,song2021mesh} selected several viewpoints to create rendered images, then applied a 2D saliency detection algorithm trained on the image saliency dataset to predict mesh saliency through 2D-to-3D projection.
Afterwards, deep learning-based saliency prediction methods \cite{zheng2019pointcloud,martin2024sal3d} were build upon the point cloud segmentation networks.
Although saliency prediction networks for non-textured meshes have emerged, there remains a lack of effective methods for handling texture and geometric features.

\noindent
\textbf{State Space Models for Vision Applications.}
The Mamba model \cite{gu2023mamba} was initially inspired by the need for efficiency in long-sequence modeling, making it particularly effective for feature extraction in visual tasks. Mamba has demonstrated strong performance across various low-level vision applications, such as image processing \cite{zhu2024vision,huang2024localmamba}, where it enhanced edge clarity and detail representation, image generation \cite{hu2024zigma,teng2024dim}, which benefited from Mamba’s capacity for intricate pattern and texture synthesis, and image restoration \cite{zheng2024u,guo2025mambair}, where it aided in the recovery of fine details from degraded inputs. It was also adapted for semantic segmentation \cite{ruan2024vm,pei2024efficientvmamba}, where its enhanced feature interactions boosted the accuracy of object detection and scene segmentation. In handling point clouds \cite{han2024mamba3d,liang2024pointmamba}, Mamba leveraged its linear complexity and powerful global modeling capabilities to efficiently manage the irregularity of point clouds, achieving improved 3D recognition.
While Mamba has been utilized in many 2D and 3D vision tasks, there remains a notable lack of research specifically addressing the integration of geometric and texture features for meshes.

\section{Dataset}
In this section, a saliency dataset of paired non-textured and textured mesh models is created using a VR eye-tracking device. By collecting eye-tracking data from observers viewing the same model under both non-textured and textured conditions, comparable saliency region distributions are obtained, highlighting the influence of texture on visual attention.

\begin{figure}[t]
  \centering
  \includegraphics[width=0.45\textwidth]{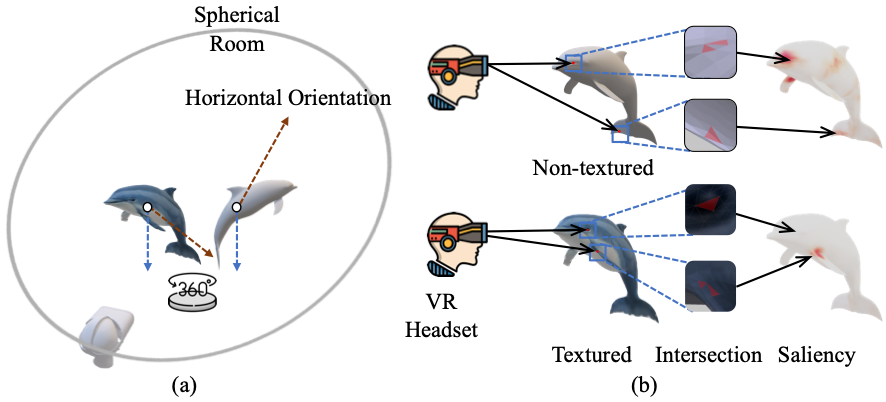}
  \setlength{\abovecaptionskip}{-12pt}
  \caption{VR eye-tracking experiment for saliency. (a) represents the virtual space setup for the experiment, (b) shows the collection of eye-tracking fixation intersections for generating saliency maps.}
\label{fig:dataset}
\end{figure}

All meshes are selected from the Free3D \cite{Free3d} open asset library for the eye-tracking experiment. This selection spans various categories, enhancing the representativeness and generalizability of the dataset, and making it applicable to a wide range of scenarios and visual applications. The Vive Pro Eye serves as the display device to present the mesh models and capture participants' eye and head movement data.
The eye-tracking experiment includes 60 participants, all of whom are newcomers to the field of computer graphics. The experimental setup for eye-tracking is illustrated in Figure \ref{fig:dataset}. To comprehensively collect eye-tracking data from various viewing angles, the models are configured for horizontal rotation, allowing participants to move freely within the space.
Throughout the experiment, the data collection program records eye-tracking data in real-time, including gaze origin, head direction, and gaze direction. By calculating the intersection points between the gaze direction and the mesh models using the Möller–Trumbore algorithm \cite{moller2005fast} accelerated by the Bounding Volume Hierarchy \cite{meister2021survey}, gaze coordinates are obtained and subsequently classified into fixation points and saccades.
Finally, the fixation points are expanded into cone-shaped beams with a Gaussian distribution (with a cone aperture of 1 degree), resulting in a smoothed visual saliency density map across the entire mesh surface.

Data analyzing leads to several findings: 1) Most textured meshes exhibit more fixation points than non-textured ones, indicating that textured models tend to attract attention more effectively. 2) The complexity of the texture significantly influences the saliency distribution of the mesh. 3) In most detailed meshes, the geometric structure continues to play a dominant role in determining saliency distribution, while the impact of texture remains relatively minor.

\begin{figure*}[t]
  \centering
  \includegraphics[width=0.92\textwidth]{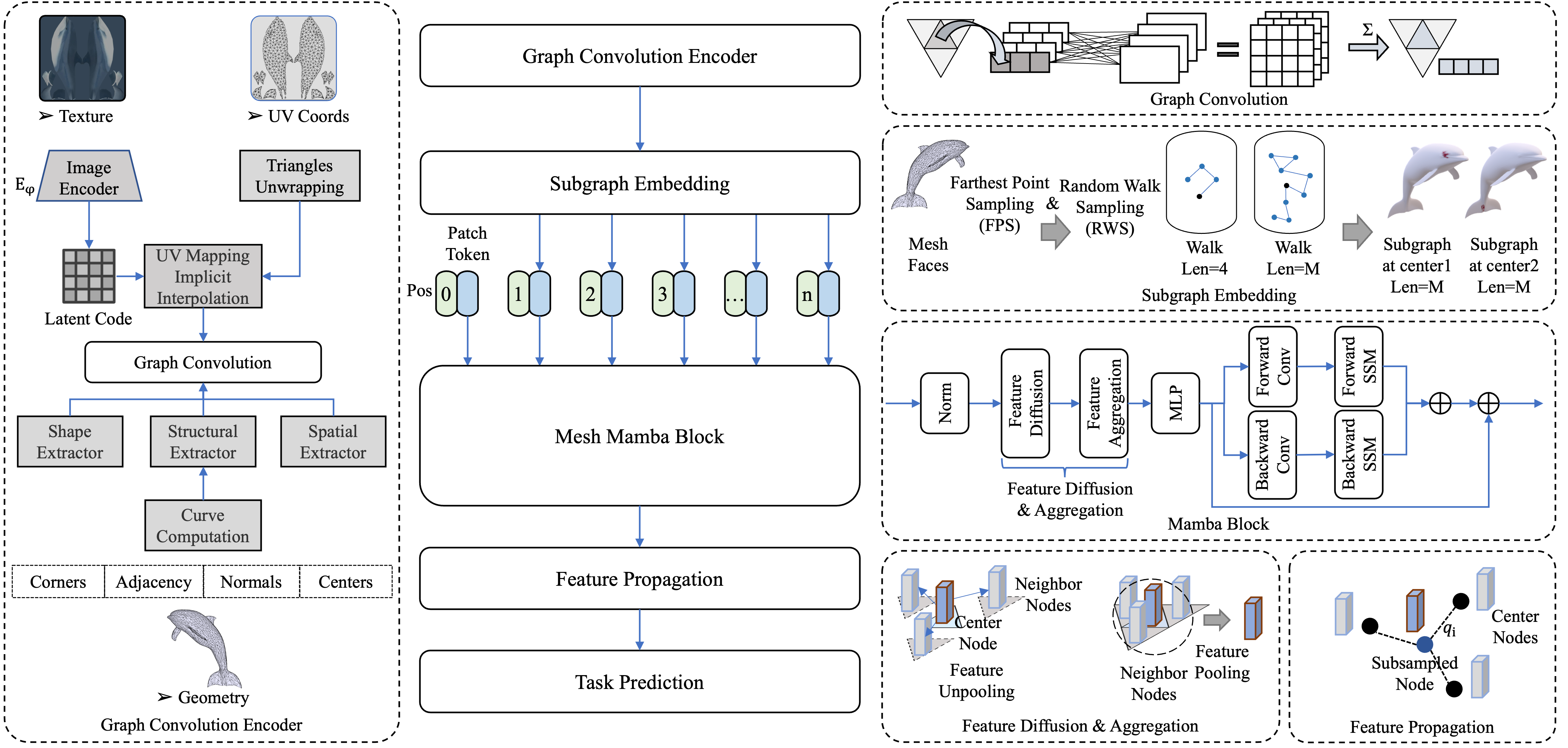}
  \setlength{\abovecaptionskip}{-0pt}
  \caption{Model architecture, including the texture alignment and geometric structure within the graph convolution encoder, along with subgraph embedding, the Mamba Block, and a feature propagation for dense prediction.}
\label{fig:model}
\end{figure*}

\section{Methodology}

In this section, we present a model designed to predict saliency maps for both non-textured and textured mesh surfaces. The saliency prediction is performed at the triangle-face level, where each face is assigned a saliency value that reflects its importance on the mesh surface.
As illustrated in Figure \ref{fig:model}, the model is structured based on the State Space Models that assigns global context modeling to the local texture and geometric features of the mesh surface, thereby enhancing the understanding of the overall structure and content. The architecture mainly comprises a graph convolution encoder, a subgraph embedding module, and a Mamba block.

\subsection{Graph Convolution Encoder}
In the context of deep learning applications involving textured meshes, it is crucial to leverage geometric information for structural description and texture information for surface feature extraction, addressing the irregularity and complexity of 3D structures.
Finally, a graph convolution block encodes the integrated local texture and geometric information into a topologically structured feature representation, utilizing the adjacency relationships of the triangular faces.

\begin{figure}[b]
  \centering
  \includegraphics[width=0.45\textwidth]{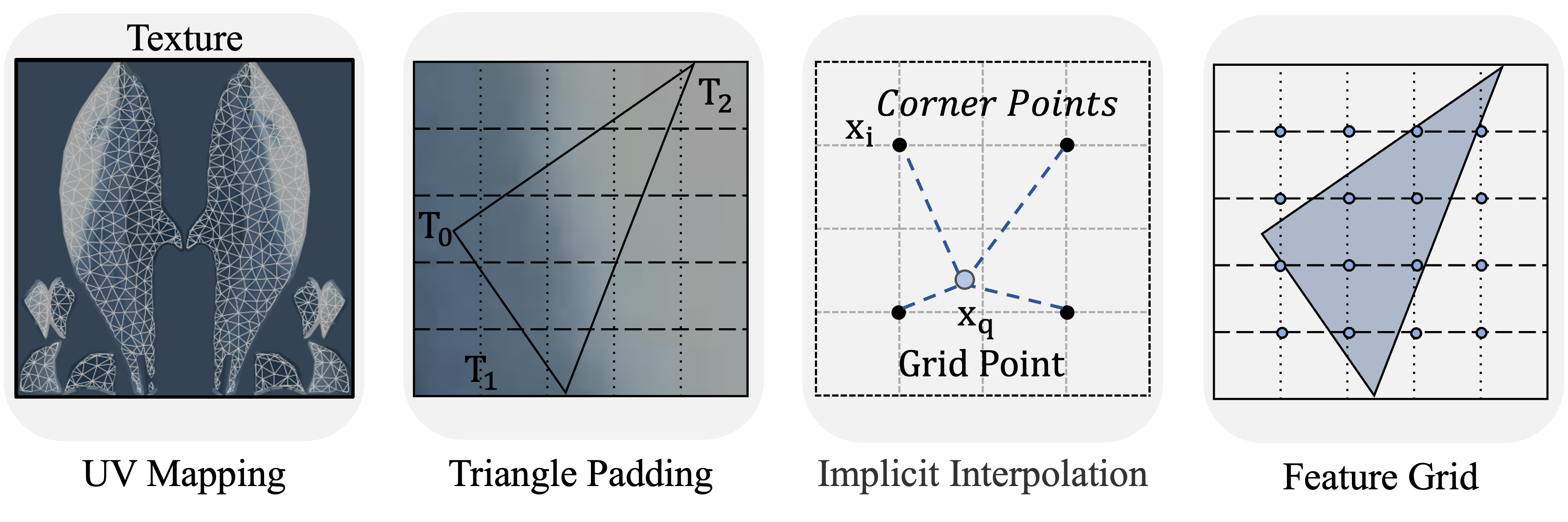}
  \setlength{\abovecaptionskip}{-12pt}
  \caption{Texture alignment with implicit representation.}
\label{fig:tex}
\end{figure}

\noindent
\textbf{Texture Alignment.}
The texture alignment retrieves texture features corresponding to each triangular from the 2D texture image based on UV mapping and accurately maps the feature vectors onto the surface of the 3D mesh.
The latent code map serves as a shared continuous implicit representation in both the texture image domain and the high-dimensional feature domain. This alignment ensures that the pixel positions corresponding to the UV coordinates in the texture image correspond precisely with those in the high-dimensional feature map.
As illustrated in Figure \ref{fig:model}, the texture image is represented by the latent code map, denoted as $E_{\varphi}(I)$, where $E$ represents the encoder, $\varphi$ refers to its parameters, and $I$ stands for the texture image. The latent code map captures feature information distributed within local regions of pixel coordinates in the image domain. By querying the latent code content using specific coordinates, it can retrieve feature values at specific locations, such as the texture feature vector corresponding to a point within a triangular face of the current textured mesh.

As illustrated in Figure \ref{fig:tex}, uniform feature sampling is conducted within the mapping range of the triangular faces to capture detailed texture features. To maintain an undistorted receptive field across all covered triangles, a consistent sampling density is ensured in both horizontal and vertical directions. Each triangle is centered, and its horizontal or vertical extent is adjusted according to its aspect ratio, filling the extended area with values from adjacent feature maps.
Furthermore, to prevent discontinuities in feature estimation, implicit interpolation is employed for feature points located at non-pixel positions. This method utilizes the values of the four nearest corners within the neighborhood. By implementing this implicit interpolation, a smooth transition of features at sub-pixel locations on the mesh is achieved, resulting in corresponding feature grid for each triangle.

\begin{figure}[b]
  \centering
  \includegraphics[width=0.32\textwidth]{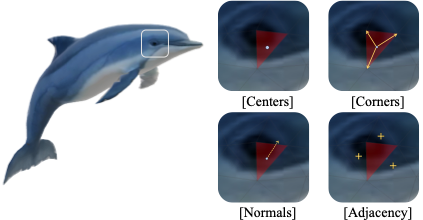}
  \setlength{\abovecaptionskip}{-0pt}
  \caption{Feature types of geometric structure.}
\label{fig:geo}
\end{figure}

\noindent
\textbf{Geometric Structure.}
In textured meshes, the geometric features encompass spatial features that encode the overall shape and spatial distribution, structural features that describe mesh topology and local geometry, and shape features that capture the irregularity and form of individual triangular faces.
In detail, spatial features are typically encoded through the relative positions $[Centers]$ between faces, reflecting the global structure of the model. Structural features reveal the local connectivity of the mesh, describing the model's topological structure and local deformations, and are used to analyze the complexity and connectivity of the model. We use the cosine similarity of $[Normals]$ between the adjacent faces $[Adjacency]$ to capture the local curve features. Shape features include the area, angles, and irregularities of triangles, aiding in the understanding of the model's local geometric properties. The angles and lengths between $[Corners]$ vectors within each face reflect the shape of the triangle. As illustrated in Figure \ref{fig:geo}, the geometric features are detailed, with specific network information available in \cite{zhang2025unified}. 

\subsection{Subgraph Embedding}

Patch embedding divides the entire structure into smaller segments and linearly maps them to vectors, facilitating the handling of local features while enhancing global understanding. In image and point cloud processing tasks, patches typically represent local regions of images or clusters of points. However, in the case of meshes, the inherent topological relationships mean that approaches like k-nearest neighbor clustering based on relative positions can disrupt adjacency connections.
Therefore, a subgraph embedding approach is implemented, as illustrated in Figure \ref{fig:model}. Initially, the farthest point sampling algorithm selects $L$ center faces from the set of triangular faces. For each center face, a random walk sampling method is employed to randomly choose neighboring triangular faces, ensuring that there are no repetitions, thus capturing a subgraph of length $M$. This subgraph, generated through random walks, retains the connectivity information of the center face while its inherent randomness introduces variability, enhancing the model's robustness against noise and uncertainty.

\subsection{Mamba Block}
As described in the original Mamba \cite{gu2023mamba}, to enhance content-aware reasoning, Mamba introduces a selection mechanism to control the propagation and interaction of information along the sequence dimension. The state propagation and sequence interaction are defined as follows:
\begin{equation}
    h_t = \bar{A}(x_t) h_{t-1} +  \bar{B}(x_t) x_t, \quad y_t = \bar{C}(x_t) h_t,
\end{equation}
where $\bar{A}(x_t)$, $\bar{B}(x_t)$, and $\bar{C}$ typically represent linear projections applied to the input $x_t$. For this discretized S4 \cite{gu2023mamba}, we utilize deep learning sequence networks for modeling.

After obtaining the subgraph patch embeddings, they are treated as a token sequence for further positional encoding. Each patch token is defined as $x_i\in\{x_1, \ldots, x_L\}$. A learnable $[cls]$ token is introduced, extending the sequence to a length of $L+1$. The sequence is then combined with learnable positional embedding $[pos]$, resulting in a sequence that enhances the model's spatial awareness capabilities:
\begin{equation}
    z_0 = [x^{cls}; x^1W;...;x^LW] + [{pos}^{cls}; {pos}^1;...;{pos}^L],
\end{equation}
where $z$ represents the output of each layer of the Mamba block, with $z_0$ forming the initial input tokens, and $W$ denotes the learnable projection matrix.

Following this, feature diffusion and aggregation operations are applied to each token, effectively enhancing the receptive field of local features. The feature diffusion operation generates $l$ pseudo-adjacent faces for each token, normalizing the feature vectors based on the mean and standard deviation before assigning them to the pseudo-adjacent faces. Subsequently, feature aggregation applies a Softmax activation to the feature vectors of both the center and pseudo-adjacent faces, followed by averaging to consolidate the neighborhood features. Through the paired diffusion and aggregation operations, the local features are appropriately enhanced.

The design of the Mamba block is illustrated in Figure \ref{fig:model}. We employ the bidirectional SSM to simultaneously process the contextual information of each element in the sequence, allowing for a more comprehensive consideration of the positions of patch tokens within the global context and their interrelationships. This approach facilitates a better understanding of the overall structure and content. Specifically, the token sequence $z_{t-1}$ is sent to the $t$-th layer of the Mamba block, and after applying a residual connection, the output $z_t$ is obtained as follows:
\begin{equation}
    z_t = SSM_{+}(f(z_{t-1})) + SSM_{-}(f(z_{t-1})) + f(z_{t-1}),
\end{equation}
where $SSM$ represents a layer of the SSM block, $f$ denotes the feature diffusion and aggregation operation, and the symbols $+$ and $-$ indicate forward and backward directions, respectively. The variable $t$ takes values in the range $\{1,...,T\}$, representing the index of $T$ Mamba blocks. 

\noindent
\textbf{Feature Propagation.}
The output from the last SSM layer retains the embedding size corresponding to the patches. To accommodate the dense prediction task for saliency, feature propagation amplifies the output embedding sequence to the appropriate scale using voting interpolation. The geometric center of the embeddings corresponds to the central face positions obtained through FPS. We weight the three nearest embedding feature values of the upsampled faces according to their distances to compute the final prediction values.

\section{Experimental Results and Discussion}

\begin{table*}[ht]
\centering
\setlength{\abovecaptionskip}{2pt}
\caption{Quantitative Results on Non-textured Mesh Saliency Prediction.}
\resizebox{0.8\textwidth}{!}{
\begin{tabular}{c|cccc|cccc|cccc}
\hline
                 & \multicolumn{4}{c|}{Geometry}                                          & \multicolumn{4}{c|}{Color}                                             & \multicolumn{4}{c}{Texture}                                            \\ \hline
Method Name      & CC $\uparrow$   & SIM $\uparrow$  & KLD $\downarrow$ & SE $\downarrow$ & CC $\uparrow$   & SIM $\uparrow$  & KLD $\downarrow$ & SE $\downarrow$ & CC $\uparrow$   & SIM $\uparrow$  & KLD $\downarrow$ & SE $\downarrow$ \\ \hline
PointNet         & 0.2626          & 0.6075          & 0.4862           & 0.0467          & 0.2449          & 0.6165          & 0.4752           & 0.0446          & 0.2813          & 0.6128          & 0.4831           & 0.0454          \\
PointNet2-SSG    & 0.3678          & 0.6452          & 0.4201           & 0.0377          & 0.3278          & 0.6447          & 0.4377           & 0.0399          & 0.3893          & 0.6502          & 0.4184           & 0.0384          \\
PointNet2-MSG    & 0.4527          & 0.6689          & 0.3846           & 0.0354          & 0.3929          & 0.6537          & 0.4086           & 0.0395          & 0.4048          & 0.6593          & 0.3983           & 0.0365          \\
PointTrans       & 0.5114          & 0.6861          & 0.3475           & 0.0314          & 0.4871          & 0.6826          & 0.3546           & 0.0338          & 0.5024          & 0.6849          & 0.3502           & 0.0333          \\
PointMixer       & 0.5104          & 0.6870          & 0.3461           & 0.0314          & 0.5137          & 0.6870          & 0.3464           & 0.0332          & 0.5131          & 0.6866          & 0.3479           & 0.0317          \\
StraTrans        & 0.5189          & 0.6875          & 0.3439           & 0.0323          & 0.4860          & 0.6792          & 0.3553           & 0.0332          & 0.4931          & 0.6789          & 0.3588           & 0.0331          \\
MeshNet          & 0.5423          & 0.7000          & 0.3390           & 0.0309          & 0.5277          & 0.6918          & 0.3551           & 0.0319          & 0.5465          & 0.7009          & 0.3408           & 0.0305          \\
MeshNet++        & 0.4978          & 0.6839          & 0.3654           & 0.0335          & 0.4285          & 0.6696          & 0.3880           & 0.0370          & 0.5096          & 0.6846          & 0.3525           & 0.0323          \\
DiffusionNet-xyz & 0.4662          & 0.6750          & 0.3787           & 0.0363          & 0.4534          & 0.6759          & 0.3839           & 0.0369          & 0.4571          & 0.6720          & 0.3934           & 0.0362          \\
DiffusionNet-hks & 0.3089          & 0.6234          & 0.4391           & 0.0421          & 0.3318          & 0.6503          & 0.4176           & 0.0391          & 0.3630          & 0.6455          & 0.4241           & 0.0388          \\
DSM\_CNN         & 0.2648          & 0.6348          & 0.4530           & 0.0431          & 0.1623          & 0.6141          & 0.5108           & 0.0494          & 0.1942          & 0.6214          & 0.4762           & 0.0459          \\
DSM\_FCN         & 0.2237          & 0.6300          & 0.4731           & 0.0444          & 0.1559          & 0.6116          & 0.4867           & 0.0482          & 0.1636          & 0.6241          & 0.4784           & 0.0465          \\
SAL3D            & 0.3988          & 0.6527          & 0.4053           & 0.0368          & 0.3775          & 0.6556          & 0.4130           & 0.0389          & 0.3316          & 0.6392          & 0.4236           & 0.0393          \\
Mamba3D          & 0.5993          & 0.7088          & 0.3345           & 0.0285          & 0.4830          & 0.6818          & 0.3566           & 0.0346          & 0.4999          & 0.6789          & 0.3655           & 0.0372          \\ \hline
Ours             & \textbf{0.6140} & \textbf{0.7171} & \textbf{0.3067}  & \textbf{0.0284} & \textbf{0.5868} & \textbf{0.7067} & \textbf{0.3104}  & \textbf{0.0296} & \textbf{0.6080} & \textbf{0.7151} & \textbf{0.3042}  & \textbf{0.0277} \\ \hline
\end{tabular}
}
\label{tab:res1}
\end{table*}

\begin{table*}[ht]
\centering
\setlength{\abovecaptionskip}{2pt}
\caption{Quantitative Results on Textured Mesh Saliency Prediction.}
\resizebox{0.8\textwidth}{!}{
\begin{tabular}{c|cccc|cccc|cccc}
\hline
                 & \multicolumn{4}{c|}{Geometry}                                          & \multicolumn{4}{c|}{Color}                                             & \multicolumn{4}{c}{Texture}                                            \\ \hline
Method Name      & CC $\uparrow$   & SIM $\uparrow$  & KLD $\downarrow$ & SE $\downarrow$ & CC $\uparrow$   & SIM $\uparrow$  & KLD $\downarrow$ & SE $\downarrow$ & CC $\uparrow$   & SIM $\uparrow$  & KLD $\downarrow$ & SE $\downarrow$ \\ \hline
PointNet         & 0.2409          & 0.6080          & 0.4963           & 0.0419          & 0.2390          & 0.6083          & 0.5046           & 0.0432          & 0.2281          & 0.6104          & 0.5097           & 0.0403          \\
PointNet2-SSG    & 0.3491          & 0.6459          & 0.4460           & 0.0362          & 0.3761          & 0.6481          & 0.4406           & 0.0375          & 0.3584          & 0.6414          & 0.4402           & 0.0372          \\
PointNet2-MSG    & 0.4749          & 0.6699          & 0.3903           & 0.0331          & 0.4280          & 0.6575          & 0.4120           & 0.0342          & 0.4389          & 0.6598          & 0.4088           & 0.0344          \\
PointTrans       & 0.5138          & 0.6821          & 0.3613           & 0.0311          & 0.4955          & 0.6756          & 0.3760           & 0.0308          & 0.5201          & 0.6817          & 0.3578           & 0.0297          \\
PointMixer       & 0.5261          & 0.6859          & 0.3531           & 0.0306          & 0.4929          & 0.6765          & 0.3755           & 0.0326          & 0.5246          & 0.6856          & 0.3576           & 0.0305          \\
StraTrans        & 0.5096          & 0.6798          & 0.3650           & 0.0304          & 0.4889          & 0.6703          & 0.3886           & 0.0315          & 0.5064          & 0.6756          & 0.3762           & 0.0314          \\
MeshNet          & 0.5512          & 0.6996          & 0.3382           & 0.0285          & 0.5535          & 0.6955          & 0.3411           & 0.0290          & 0.5605          & 0.7002          & 0.3371           & 0.0286          \\
MeshNet++        & 0.4167          & 0.6623          & 0.4104           & 0.0366          & 0.4389          & 0.6642          & 0.4064           & 0.0343          & 0.4183          & 0.6635          & 0.4012           & 0.0367          \\
DiffusionNet-xyz & 0.4351          & 0.6660          & 0.4219           & 0.0356          & 0.4198          & 0.6623          & 0.4226           & 0.0361          & 0.4296          & 0.6615          & 0.4222           & 0.0344          \\
DiffusionNet-hks & 0.3293          & 0.6065          & 0.4569           & 0.0397          & 0.3496          & 0.6333          & 0.4452           & 0.0388          & 0.3781          & 0.6393          & 0.4374           & 0.0365          \\
DSM\_CNN         & 0.2174          & 0.6186          & 0.4902           & 0.0408          & 0.2082          & 0.6108          & 0.4984           & 0.0422          & 0.2189          & 0.6022          & 0.5186           & 0.0465          \\
DSM\_FCN         & 0.2040          & 0.6160          & 0.5008           & 0.0414          & 0.2031          & 0.6096          & 0.5073           & 0.0450          & 0.2163          & 0.6038          & 0.5039           & 0.0445          \\
SAL3D            & 0.4446          & 0.6674          & 0.4008           & 0.0332          & 0.4206          & 0.6560          & 0.4297           & 0.0363          & 0.3588          & 0.6405          & 0.4562           & 0.0374          \\
Mamba3D          & 0.5835          & 0.7011          & 0.3477           & 0.0284          & 0.4574          & 0.6633          & 0.3983           & 0.0357          & 0.5013          & 0.6769          & 0.3797           & 0.0342          \\ \hline
Ours             & \textbf{0.6066} & \textbf{0.7113} & \textbf{0.3134}  & \textbf{0.0267} & \textbf{0.5957} & \textbf{0.7105} & \textbf{0.3252}  & \textbf{0.0281} & \textbf{0.6305} & \textbf{0.7232} & \textbf{0.2888}  & \textbf{0.0265} \\ \hline
\end{tabular}
}
\label{tab:res2}
\end{table*}

Current mesh saliency prediction methods are typically designed for non-textured meshes, focusing primarily on geometric features. The objective here is to extend saliency prediction to both the textured and non-textured versions of the same mesh by integrating geometric and texture information. To achieve this, saliency prediction and evaluation are conducted under both textured and non-textured conditions for the same mesh.

\noindent
\textbf{Experimental Setup.}
The evaluation follows a progressive approach, beginning with an assessment of different models using only geometric structure. To ensure fairness, our model extracts only geometric features and is compared with other models under the same conditions. Next, we introduce vertex colors to enhance our model with additional contextual information. Notably, we apply the same color extraction module to all the comparison methods to ensure a fair comparison. Finally, we incorporate the texture module to improve the realism and accuracy of all methods when texture features are included. Similarly, all methods use the same texture alignment module for a uniform evaluation. Through extensive cross validations, we assess the impact of geometric and texture features on the predictive capabilities of various methods.

\begin{figure*}[t]
  \centering
  \includegraphics[width=0.88\textwidth]{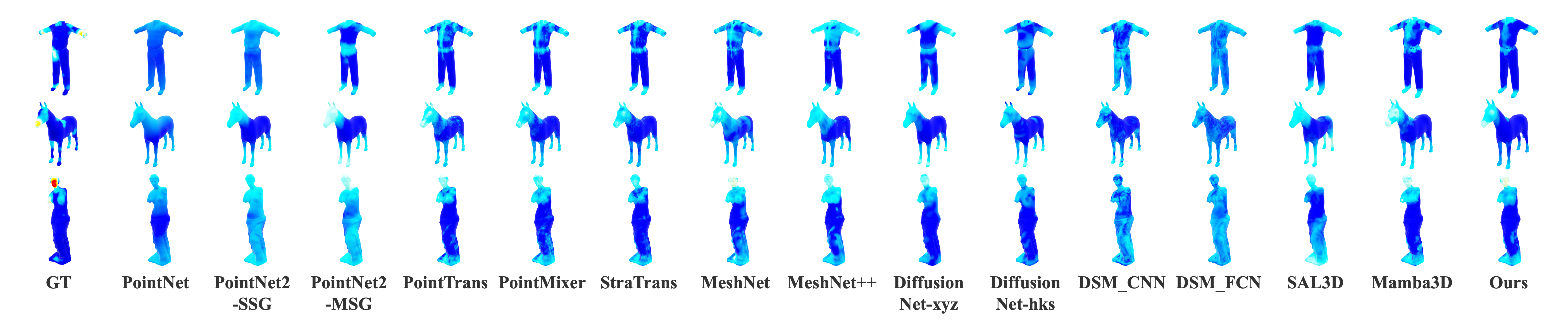}
  \setlength{\abovecaptionskip}{-0pt}
  \caption{Visualization results of compared methods on the non-textured meshes.}
\label{fig:res1}
\end{figure*}
\begin{figure*}[t]
  \centering
  \includegraphics[width=0.88\textwidth]{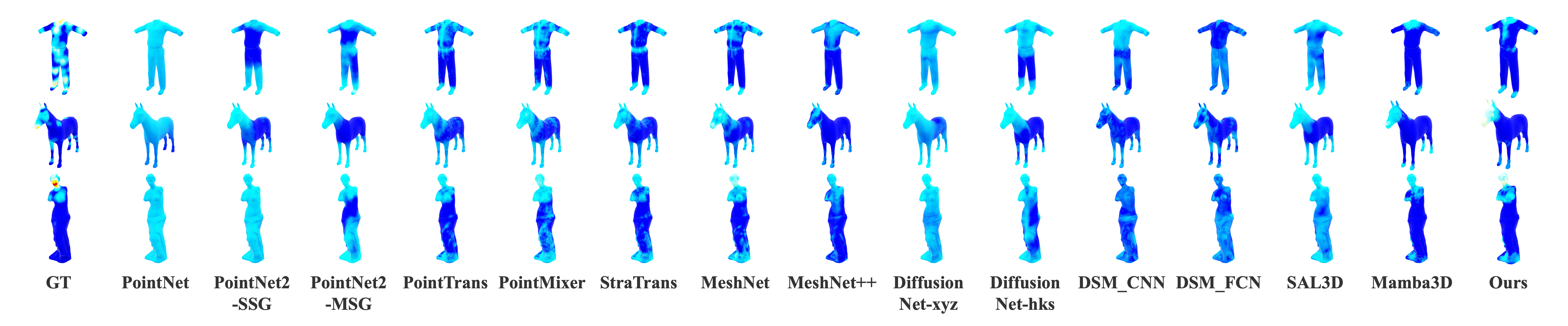}
  \setlength{\abovecaptionskip}{-0pt}
  \caption{Visualization results of compared methods on the textured meshes.}
\label{fig:res2}
\end{figure*}

\noindent
\textbf{Implementation Details.}
We utilize standard evaluation metrics for saliency prediction, including Correlation Coefficient (CC), Similarity (SIM), Kullback-Leibler Divergence (KLD), and Saliency Error (SE) which is measured using Mean Squared Error (MSE). The training process is conducted on a server equipped with dual NVIDIA GeForce RTX 4090 GPUs and an Intel i9 processor. We employ the AdamW optimizer with L1 loss as the loss function. The initial learning rate is set to 1e-3 and decreases by a factor of 0.1 every 50 epochs, with training running for a total of 150 epochs.
The dataset is split into 80\% for training and 20\% for testing. Each mesh includes two types of ground truth saliency maps, obtained under both textured and non-textured conditions.

\noindent
\textbf{Baselines.}
The state-of-the-art deep models for mesh saliency prediction, DSM \cite{nousias2023deep} (which incorporates both CNN-based and FCN-based approaches for 3D saliency patch descriptors) and SAL3D \cite{martin2024sal3d}, are designed for non-textured meshes and are employed as baseline methods for performance comparison.
Due to the limited availability of saliency prediction models for comparative analysis, several 3D segmentation models are adapted by modifying the output layers to accommodate the saliency prediction task, ensuring a fair evaluation. Since the center coordinates of mesh faces can be represented as point cloud data, point-based models are also incorporated for saliency prediction. These include PointNet \cite{qi2017pointnet}, PointNet2 \cite{qi2017pointnet++}, PointTrans \cite{zhao2021point}, PointMixer \cite{choe2022pointmixer}, StraTrans \cite{lai2022stratified}, and Mamba3D \cite{han2024mamba3d}.
In addition to point-based models, mesh-based architectures are also adapted to the saliency prediction task to maintain methodological consistency across different approaches. The selected mesh-based models include MeshNet \cite{feng2019meshnet}, MeshNet++ \cite{singh2021meshnet++}, and DiffusionNet \cite{sharp2022diffusionnet} (where the xyz and hks variants correspond to networks utilizing positional data and heat kernel signatures as input).


\subsection{Performance Evaluation on Proposed Dataset}
\subsubsection{Saliency Prediction for Non-textured Mesh}

\noindent
\textbf{Geometry.}
We evaluate the performance of various methods on non-textured saliency prediction based solely on geometric structure. The quantitative results are shown in the Geometry column of Table \ref{tab:res1}. Methods adapted from PointNet exhibit relatively poor performance, while more advanced approaches based on Transformers and Mamba3D achieve superior results. Additionally, methods that integrate structural features, such as MeshNet, outperform most other methods. In contrast, the DSM method, which is trained on patch-scale regions of the mesh surface, performs poorly. The latest SAL3D method, based on PointNet2, also shows a significant performance gap. Furthermore, as shown in Figure \ref{fig:res1}, the visualized results of our model display clear boundaries and effectively cover the main salient regions.

\noindent
\textbf{Color.}
To consistently assess the impact of vertex colors on non-textured saliency prediction, all methods integrate the same color extraction module. Vertex colors are mapped to vertex coordinates using UV mapping from the texture image. As indicated in the Color column of Table \ref{tab:res1}, integrating vertex colors into our model leads to a noticeable performance decline. Similarly, most methods also exhibit significant performance degradation, indicating that vertex color has limited capability in enhancing feature representations for non-textured mesh saliency prediction.

\noindent
\textbf{Texture.}
To further explore the influence of finer texture features on prediction, the same texture module is incorporated into all methods. As shown in the Texture column of Table \ref{tab:res1}, incorporating the texture alignment module does not improve the performance of our model. Similarly, for most methods, texture features are redundant for non-textured mesh saliency prediction and may even introduce noise.

\subsubsection{Saliency Prediction for Textured Mesh}

\noindent
\textbf{Geometry.}
In saliency prediction for textured mesh, we begin by evaluating the performance of various methods based solely on geometric structure. The quantitative results are shown in the Geometry column of Table \ref{tab:res2}, where the performance aligns with the previous evaluation. Our model continues to exhibit the highest performance.

\noindent
\textbf{Color.}
The results after incorporating the same color extraction module across all methods are presented in the Color column of Table \ref{tab:res2}. Integrating vertex colors into the models results in a noticeable performance decline for most methods. This indicates that vertex color still has limited feature extraction capabilities for textured meshes, which exhibit complex visual effects due to varying texture patterns. Vertex colors are insufficient to represent the complex visual cues in textured meshes.

\noindent
\textbf{Texture.}
In the final step, the texture alignment module is applied to all methods, and the results in the Texture column of Table \ref{tab:res2} demonstrate that the integration of texture features significantly improves the performance of most models. As seen in the saliency prediction visualization results in Figure \ref{fig:res2}, our model demonstrates clear boundaries and effectively covers the main salient regions. Other methods that had already performed well with only geometric features also show some improvement. These results reflect the enhanced performance in saliency prediction under textured conditions through the combination of geometric structures and texture features.

\subsubsection{Non-textured vs Textured Saliency Analysis}
Through cross validations of saliency prediction for non-textured and textured meshes, several key conclusions can be drawn. For non-textured meshes, geometric structure alone as the sole visual feature is sufficient to reconstruct an accurate saliency distribution. The addition of vertex colors and texture information, however, introduces redundancy that can degrade model performance, though it does not lead to catastrophic forgetting, demonstrating the model's robustness. For textured meshes, the integration of texture and geometric features significantly improves model performance. However, the insufficient capacity of vertex colors to represent complex textures ultimately hampers the model's overall effectiveness.

\begin{table}[htbp]
\centering
\setlength{\abovecaptionskip}{2pt}
\caption{Quantitative Results on SAL3D Dataset.}
\resizebox{0.35\textwidth}{!}
{
\begin{tabular}{c|cccc}
\hline
Method Name      & CC $\uparrow$   & SIM $\uparrow$  & KLD $\downarrow$ & SE $\downarrow$ \\ \hline
PointNet         & 0.5351          & 0.6809          & 0.3526           & 0.0215          \\
PointNet2-SSG    & 0.5797          & 0.7086          & 0.2879           & 0.0188          \\
PointNet2-MSG    & 0.6190          & 0.7188          & 0.2743           & 0.0187          \\
PointTrans       & 0.4663          & 0.6706          & 0.3624           & 0.0230          \\
PointMixer       & 0.4670          & 0.6697          & 0.3602           & 0.0226          \\
StraTrans        & 0.4507          & 0.6661          & 0.3704           & 0.0228          \\
MeshNet          & 0.5904          & 0.7060          & 0.2947           & 0.0175          \\
MeshNet++        & 0.6513          & 0.7243          & 0.2556           & 0.0173          \\
DiffusionNet-xyz & 0.4144          & 0.6607          & 0.3800           & 0.0217          \\
DiffusionNet-hks & 0.5371          & 0.6860          & 0.3422           & 0.0198          \\
DSM\_CNN         & 0.4010          & 0.6277          & 0.4068           & 0.0212          \\
DSM\_FCN         & 0.3744          & 0.6569          & 0.4000           & 0.0238          \\
SAL3D            & 0.5736          & 0.7016          & 0.3104           & 0.0209          \\
Mamba3D          & 0.6181          & 0.7203          & 0.2643           & 0.0179          \\ \hline
Ours             & \textbf{0.6763} & \textbf{0.7342} & \textbf{0.2303}  & \textbf{0.0170} \\ \hline
\end{tabular}
}
\label{tab:res3}
\end{table}

\subsection{Performance Evaluation on SAL3D Dataset}
To further validate the proposed model, we conduct experiments on the SAL3D \cite{martin2024sal3d} dataset, which contains 58 non-textured meshes. We use 46 meshes for training and 12 for testing. As shown in Table \ref{tab:res3}, the simpler geometric structures in this dataset lead to overall higher performance across all methods. Notably, structure-focused methods like MeshNet achieve particularly strong results. Our model also demonstrates superior performance on the SAL3D dataset.

\begin{figure}[]
  \centering
  \includegraphics[width=0.45\textwidth]{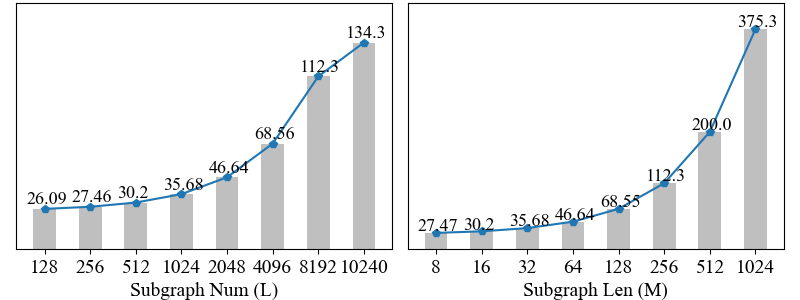}
  \setlength{\abovecaptionskip}{-8pt}
  \caption{Model's FLOPs increase linearly with subgraph number and length.}
\label{fig:flops}
\end{figure}

\begin{table}[]
\centering
\setlength{\abovecaptionskip}{0pt}
\caption{The Performance of Different Components.}
\resizebox{0.38\textwidth}{!}{
\begin{tabular}{l|cccc}
\hline
\# Subject       & CC $\uparrow$ & SIM $\uparrow$ & KLD $\downarrow$ & SE $\downarrow$ \\ \hline
w/o Texture      & 0.6066        & 0.7113         & 0.3134           & 0.0267          \\
w/o Spatial      & 0.6175        & 0.7161         & 0.2931           & 0.0278          \\
w/o Shape        & 0.5403        & 0.6903         & 0.3634           & 0.0324          \\
w/o Curve        & 0.6036        & 0.7160         & 0.3058           & 0.0303          \\ \hline
w/  Backbone-T     & 0.6113        & 0.7204         & 0.2975           & 0.0270          \\
w/o Graph Conv & 0.5993        & 0.7159         & 0.2970           & 0.0285          \\
w/o Subgraph     & 0.6237        & 0.7208         & 0.3048           & 0.0298          \\
w/o Feature D\&A & 0.5889        & 0.7106         & 0.3123           & 0.0294          \\
w/o SSM-         & 0.6203        & 0.7186         & 0.2935           & 0.0272          \\
w/o SSM+         & 0.6199        & 0.7164         & 0.2947           & 0.0275          \\ \hline
\end{tabular}
}
\label{tab:abs1}
\end{table}

\subsection{Ablation Study}
In Table \ref{tab:abs1}, we conduct several ablation studies on our model under textured conditions, demonstrating the necessity of each component in the model architecture by removing them individually. Initially, we examine the influence of input geometric and texture feature types on the final outcome. The experiments show that the shape features of triangular faces have the greatest impact on the surface saliency distribution. Next, we replace the backbone with a Transformer and remove the graph convolution layer. We then substitute the RWS of subgraph embedding with KNN clustering centered on FPS points. Subsequently, we individually remove key components within the Mamba block, including feature diffusion and aggregation, forward SSM layer and backward SSM layer. The experimental results demonstrate the effectiveness of the feature types and model structure.
Finally, as shown in Figure \ref{fig:flops}, we present the computational FLOPs for varying subgraph counts and lengths, demonstrating a linear growth trend as these parameters increase.

\begin{figure}[!t]
  \centering
  \includegraphics[width=0.45\textwidth]{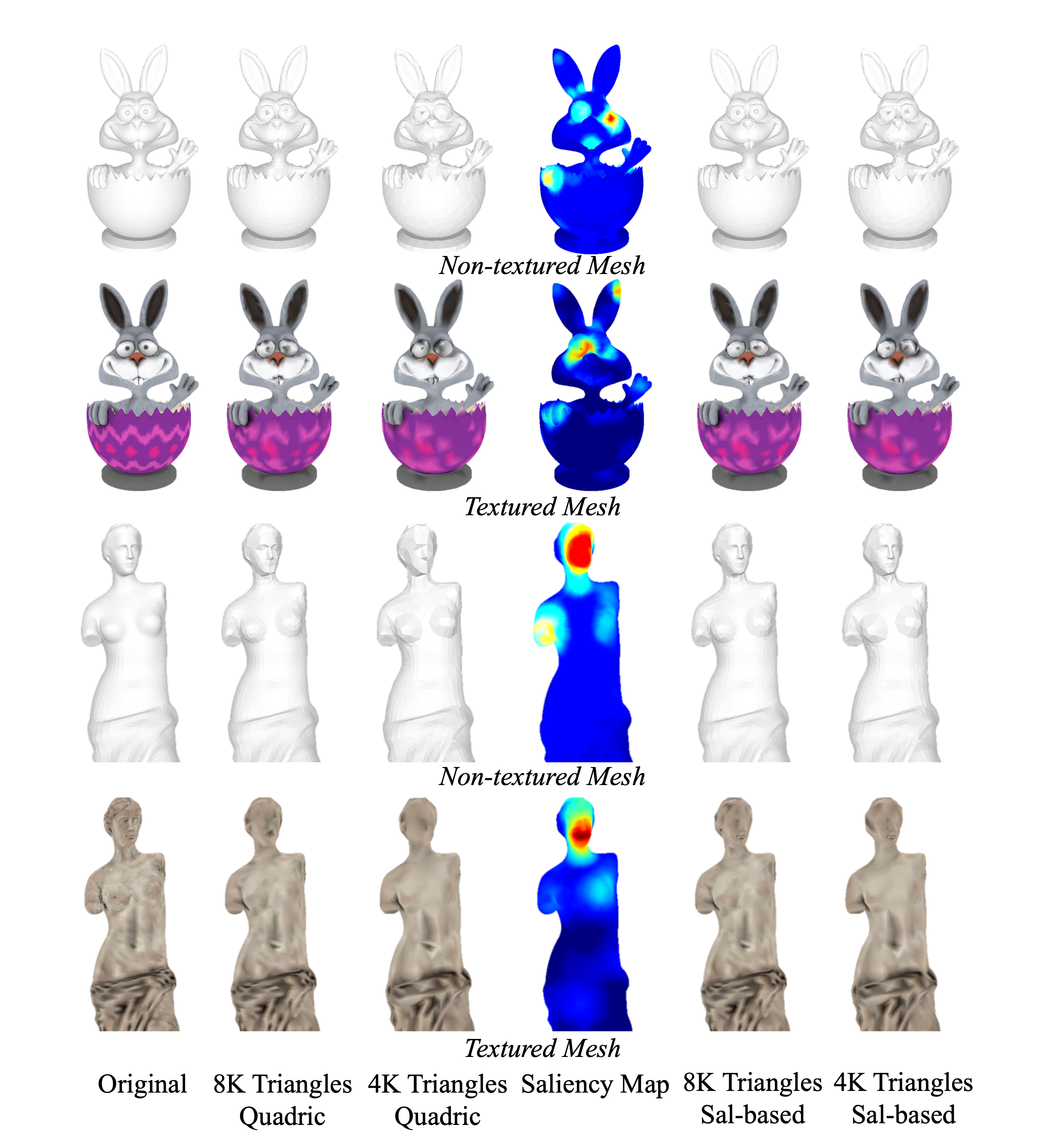}
  \setlength{\abovecaptionskip}{-4pt}
  \caption{Mesh simplification results with quadric-based algorithm and saliency-based simplification.}
\label{fig:simplify}
\end{figure}

\subsection{Discussion on Saliency-Based Simplification}
When observing 3D objects, the human eye tends to focus on prominent features, such as unique colors or specific shapes. Certain semantically meaningful regions, like faces, ornate furniture in a room, or door handles, are particularly likely to draw visual focus. These key areas in mesh models cannot be fully captured through abstracted shape and color features alone, but saliency detection offers a way to identify them.
In this block, we modify the classic quadric-based simplification algorithm \cite{garland1997surface} by incorporating saliency-based guidance, and compare the results with the original algorithm. As demonstrated in Figure \ref{fig:simplify}, we present two examples with 8K and 4K triangle counts. The results clearly show that areas with high visual attention, such as the face, eyes, and torso, retain significantly more detail in both non-textured and textured meshes.

\section{Conclusion}
This paper addresses the saliency prediction task for both non-textured and textured meshes by proposing a unified architecture based on state space model, which enables reliable saliency prediction across varying texture conditions.
By integrating geometric and texture features, the model achieves a comprehensive representation of the mesh. The SSM further augments the contextual awareness of local features, while the bidirectional state propagation mechanism captures intricate spatial relationships, thereby enhancing the model’s ability to interpret complex structural interactions.
The experimental results demonstrate that although geometric features alone are effective for non-textured saliency prediction, incorporating texture information notably improves performance for textured meshes, especially in complex visual scenes. This research highlights the crucial role of multimodal feature integration in achieving accurate 3D saliency predictions and establishes a valuable benchmark for advancing future studies in this area.

\section{Acknowledgement}
This work was supported in part by the National Natural Science Foundation of China under Grant 62271312, Grant 6213000376, Grant 62101325, Grant 62101326, and Grant 62377011.

{
    \small
    \bibliographystyle{ieeenat_fullname}
    \bibliography{main}
}


\end{document}